\documentclass[10pt,twocolumn,letterpaper]{article}

\usepackage{iccv}
\usepackage{times}
\usepackage{epsfig}
\usepackage{graphicx}
\usepackage{amsmath}
\usepackage{amssymb}
\usepackage{multirow}
\usepackage{adjustbox}
\usepackage{textcomp}
\usepackage{enumitem}
\usepackage{lipsum}
\newcommand\blfootnote[1]{%
  \begingroup
  \renewcommand\thefootnote{}\footnote{#1}%
  \addtocounter{footnote}{-1}%
  \endgroup
}

\renewcommand{\textrightarrow}{$\rightarrow$}

\usepackage[pagebackref=true,breaklinks=true,letterpaper=true,colorlinks,bookmarks=false]{hyperref}

\iccvfinalcopy 

\def\iccvPaperID{3636} 

\ificcvfinal\fi
\begin{document}

\title{Unsupervised Video Interpolation Using Cycle Consistency}

\author{Fitsum A. Reda \quad Deqing Sun$^{*}$ \quad Aysegul Dundar \quad Mohammad Shoeybi \quad Guilin Liu\\ Kevin J. Shih \quad Andrew Tao \quad Jan Kautz \quad Bryan Catanzaro\\
NVIDIA} 

\maketitle
\thispagestyle{empty}

\begin{abstract}

Learning to synthesize high frame rate videos via interpolation requires large quantities of high frame rate training videos, which, however, are scarce, especially at high resolutions. Here, we propose unsupervised techniques to synthesize high frame rate videos directly from low frame rate videos using cycle consistency. For a triplet of consecutive frames, we optimize models to minimize the discrepancy between the center frame and its cycle reconstruction, obtained by interpolating back from interpolated intermediate frames. This simple unsupervised constraint alone achieves results comparable with supervision using the ground truth intermediate frames. We further introduce a pseudo supervised loss term that enforces the interpolated frames to be consistent with predictions of a pre-trained interpolation model. The pseudo supervised loss term, used together with cycle consistency, can effectively adapt a pre-trained model to a new target domain. With no additional data and in a completely unsupervised fashion, our techniques significantly improve pre-trained models on new target domains, increasing PSNR values from 32.84dB to 33.05dB on the Slowflow and from 31.82dB to 32.53dB on the Sintel evaluation datasets. Code is available at \textcolor{magenta}{\href{https://github.com/NVIDIA/unsupervised-video-interpolation}{https://github.com/NVIDIA/unsupervised-video-interpolation}.} \blfootnote{$^*$Currently affiliated with Google.}

\end{abstract}

\section{Introduction}

 \begin{figure}[t]
    \centering
    \includegraphics[trim={0 0 0 0},clip,width=1.0\linewidth]{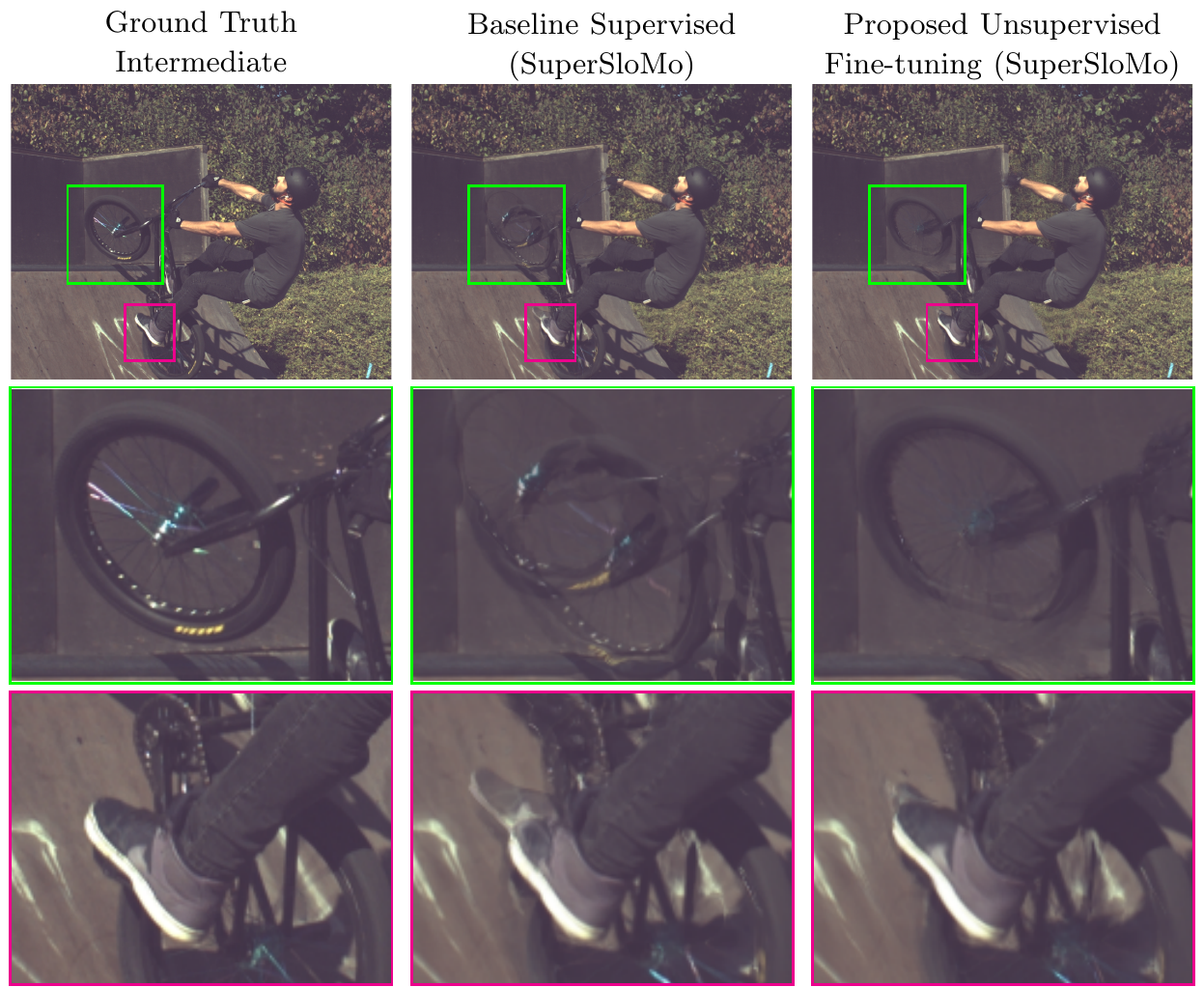}
    \caption{Visual results of a sample from Slowflow dataset. Baseline supervised model is trained with Adobe and YouTube datasets and proposed unsupervised model is fine-tuned with Slowflow. See project \textcolor{magenta}{\href{https://nv-adlr.github.io/publication/2019-UnsupervisedVideoInterpolation}{website}} for video comparisons.}
    \label{fig:Front_page_figure_slowflow}
\end{figure}

With the advancement of modern technology, consumer-grade smartphones and digital cameras can now record videos at high frame rates (e.g. 240 frames-per-second). However, achieving this comes at the cost of high power consumption, larger storage requirements, and reduced video resolution. Given these limitations, regular events are not typically recorded at high frame rates. Yet, important life events happen unexpectedly and hence tend to be recorded at standard frame rates.
It is thus greatly desirable to have the ability to produce arbitrarily high FPS videos from low FPS videos. 

Video frame interpolation addresses this need by generating one or more intermediate frames from two consecutive frames.
Increasing the number of frames in videos essentially allows one to visualize events in slow motion and appreciate content better.
Often, video interpolation techniques are employed to increase the frame rate of already recorded videos, or in streaming applications to provide a high refresh rate or a smooth viewing experience. 

Video interpolation is a classical vision and graphics problem~\cite{Baker2009OcclusionInterpolation,szeliski1999prediction,zitnick2004high}  and has recently received renewed research attention~\cite{long16learning,Niklaus_2017_ICCV,long2016learning,Niklaus_2018_CVPR}. Particularly, supervised learning with convolutional neural networks (CNNs) has been widely employed to learn video interpolation from paired input and ground truth frames, often collected from raw video data. For instance, recent CNN-based approaches such as~\cite{jiang2018super} and~\cite{Niklaus_2017_ICCV}, trained with large quantities of public high FPS videos, obtain high quality interpolation results when the test videos are similar to the training ones.

However, these methods may fail if the training data differ from the target domain. For instance, the target domain might be to slow down videos of fish taken underwater, but available training data only contains regular outdoor scenes, thus leading to a content gap. Additionally, there might be more subtle domain gaps due to differences such as camera parameters, encoding codecs, and lighting, leading to the well-known co-variate shift problem \cite{shimodaira2000improving}. It is impractical to address the issue by collecting high FPS videos covering all possible scenarios, because it is expensive to capture and store very high FPS videos, e.g., videos with more than 1000-fps at high spatial resolutions.

In this work, we propose a set of unsupervised learning techniques to alleviate the need for high FPS training videos and to shrink domain gaps in video interpolation. Specifically, we propose to learn video frame interpolation, without paired training data, by enforcing models to satisfy a cycle consistency  constraint \cite{brislin1970back} in the time. That is, for a given triplet of consecutive frames, if we generate the two intermediate frames between the two consecutive frames, and generate back their intermediate frame, the resulting frame must match the original input middle frame (shown schematically in Fig.~\ref{fig:cycle_consistency_figure}). We show such simple constraint alone is effective to learn video interpolation, and achieve results that compete with supervised approaches. 

\begin{figure}[t]
\centering
\includegraphics[trim={30 160 30 90},clip,width=1.0\linewidth]{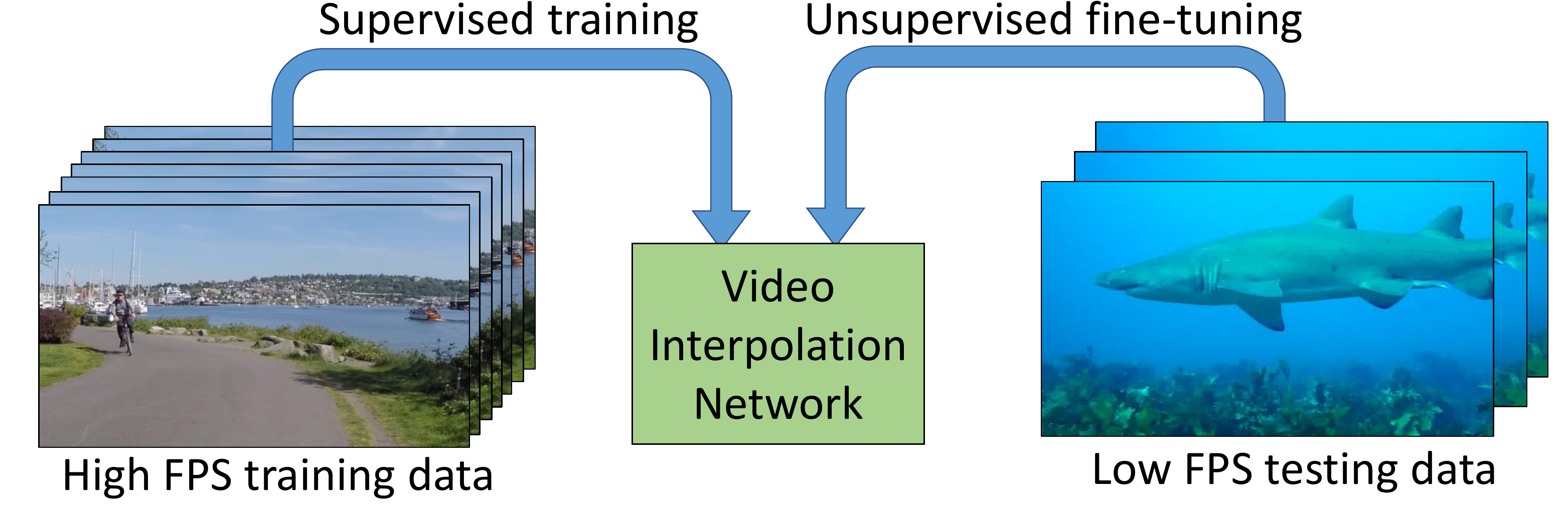}
\caption{Video interpolation methods may fail if the training data differ from the test data. In this work, we propose an unsupervised fine-tuning technique to reduce domain gaps.}
\label{fig:teaser}
\end{figure}

In domain adaptation applications, where we have access to models pre-trained on out-of-domain datasets, but lack ground truth frames in target domains, we also propose unsupervised fine-tuning techniques that leverage such pre-trained models (See Fig.~\ref{fig:teaser}). We fine-tune models on target videos, with no additional data, by optimizing to jointly satisfy cycle consistency and minimize the discrepancy between generated intermediate frames and corresponding predictions from the known pre-trained model. We demonstrate our joint optimization strategy leads to significantly superior accuracy in upscaling frame rate of target videos than fine-tuning with cycle consistency alone or directly applying the pre-trained models on target videos (see Fig. \ref{fig:Front_page_figure_slowflow}). 

Cycle consistency has been utilized for image matching \cite{zhou2015multi}, establishing dense 3D correspondence over object instances \cite{zhou2015flowweb}, or in learning unpaired image-to-image translation in conjunction with Generative Adversarial Networks (GANs) \cite{zhu2017unpaired}. To the best of our knowledge, this is the first attempt to use a cycle consistency constraint to learn video interpolation in a completely unsupervised way. 

To summarize, the contributions of our work include:
\begin{itemize}[topsep=0pt,noitemsep]
    \item We propose unsupervised approaches to learn video interpolation in the absence of paired training data by optimizing neural networks to satisfy cycle consistency constraints in time domain.
    
    \item We learn to synthesize arbitrarily high frame rate videos by learning from only low frame rate raw videos.
    
    \item We demonstrate the effectiveness of our unsupervised techniques in reducing domain gaps in video interpolation.

\end{itemize}


\section{Related Works}

\noindent\textbf{Video Interpolation:} The task is to interpolate intermediate frames between a pair of input frames. Classical approaches such as \cite{herbst2009occlusion} and \cite{mahajan2009moving} estimate optical flow and interpolate intermediate frames at intermediate positions along the estimated trajectory of pixels, and further make use of forward and backward optical flow consistency to reason about occlusions. 

Given the recent rise in popularity of deep learning methods, several end-to-end trainable methods have been proposed for video interpolation. Specifically, these methods can be trained to interpolate frames using just input and target frames and no additional supervision. Liu et al.~\cite{liu2017video} and Jiang  et al.~\cite{jiang2018super} both indirectly learn to predict optical flow using frame interpolation. Works such as ~\cite{Niklaus_2017_CVPR,Niklaus_2017_ICCV} are similarly end-to-end trainable, but instead of learning optical flow vectors to warp pixels, they predict adaptive convolutional kernels to apply at each location of the two input frames. Our work presents unsupervised techniques to train or fine-tune any video interpolation model, for instance the Super SloMo~\cite{jiang2018super}, which predicts multiple intermediate frames, or the Deep Voxel Flow~\cite{liu2017video}, which predicts one intermediate frame. 

\noindent\textbf{Cycle Consistency:}
One of the key elements of our proposed method is the use of a cycle consistency constraint. This constraint encourages the transformations predicted by a model to be invertible, and is often used to regularize the model behavior when direct supervision is unavailable. Cycle consistency has been used in a variety of applications, including determining the quality of language translations~\cite{brislin1970back}, semi-supervised training for image-description generation~\cite{mao2016generation}, dense image correspondences~\cite{zhou2016learning}, identifying false visual relations in structure from motion~\cite{zach2010disambiguating}, and image-to-image translation~\cite{zhu2017unpaired}, to name a few.

A cycle consistency constraint, in the context of video interpolation, means that we should be able to reconstruct the original input frames by interpolating between predicted intermediate frames at the appropriate time stamps. Most related to our work is~\cite{liu2019deep}, which uses such a constraint to regularize a fully supervised video interpolation model. Our work differs in several critical aspects. First, our method is based on the Super SloMo~\cite{jiang2018super} architecture, and is thus capable of predicting intermediate frames at arbitrary timestamps, whereas ~\cite{liu2019deep} is specifically trained to predict the middle timestamp. Next, and most critically, our proposed method is fully unsupervised. This means that the target intermediate frame is never used for supervision, and that it can learn to produce high frame rate interpolated sequences from any lower frame rate sequence.

\section{Method}\label{Methodology}
In this work, we propose to learn to interpolate arbitrarily many intermediate frames from a pair of input frames, in an unsupervised fashion, with no paired intermediate ground truth frames. Specifically, given a pair of input frames $\textbf{I}_{0}$ and $\textbf{I}_{1}$, we generate intermediate frame $\hat{\textbf{I}}_{t}$, as %
\begin{equation} \label{eq:1}
\hat{\textbf{I}}_{t} = \mathcal{M}\big(\textbf{I}_{0},\textbf{I}_{1},t\big), \end{equation} 
where $t\in(0,1)$ is time, and $\mathcal{M}$ is a video frame interpolation model we want to learn without supervision. We realize $\mathcal{M}$ using deep convolutional neural networks (CNN). We chose CNNs as they are able to model highly non-linear mappings, are easy to implement, and have been proven to be robust for various vision tasks, including image classification, segmentation, and video interpolation.

Inspired by the recent success in learning unpaired image-to-image translation using Generative Adversarial Networks (GAN)  \cite{zhu2017unpaired}, we propose to optimize $\mathcal{M}$ to maintain cycle consistency in time. Let  $\textbf{I}_{0}$, $\textbf{I}_{1}$ and $\textbf{I}_{2}$ are a triplet of consecutive input frames. We define the time-domain cycle consistency constraint such that for generated intermediate frames at time $t$ between $(\textbf{I}_{0}, \textbf{I}_{1})$ and between $(\textbf{I}_{1}, \textbf{I}_{2})$, a subsequently generated intermediate frame at time $(1-t)$ between the interpolated results $(\hat{\textbf{I}}_{t}, \hat{\textbf{I}}_{t+1})$ must match the original middle input frame $\textbf{I}_{1}$. Mathematically, a cycle reconstructed frame using $\mathcal{M}$ is given by, 
\begin{equation} \label{eq:2}
\hat{\textbf{I}}_{1} = \mathcal{M}\Big(\mathcal{M}\big(\textbf{I}_{0},\textbf{I}_{1},t\big),  \mathcal{M}\big(\textbf{I}_{1},\textbf{I}_{2}, t\big), 1-t\Big).
\end{equation}

\begin{figure}[t]
\centering
\includegraphics[trim={140 180 180 40},clip,width=1.0\linewidth]{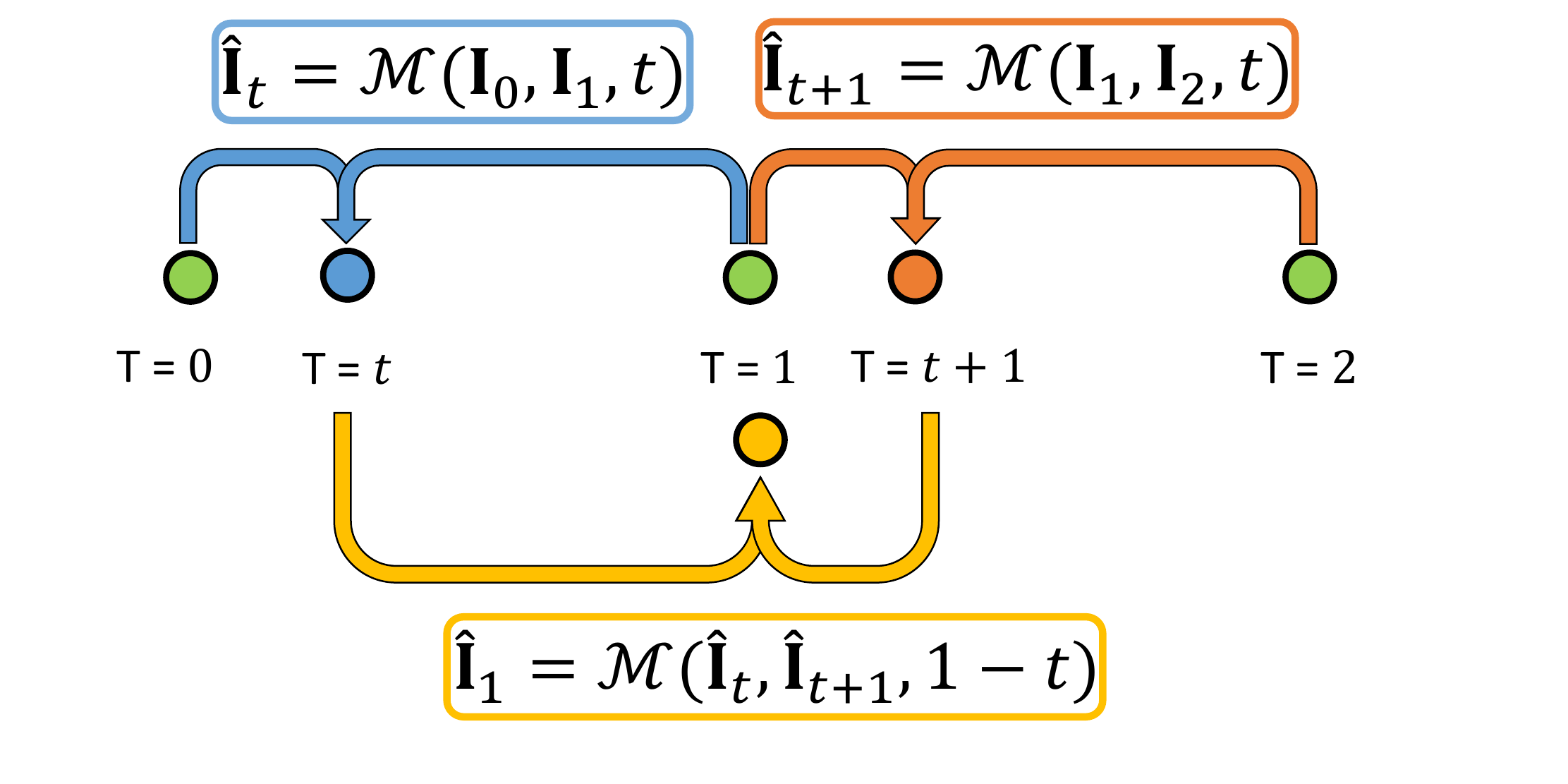}
\caption{An overview of time-domain cycle consistency constrain. $\textbf{I}_{0}$, $\textbf{I}_{1}$ and $\textbf{I}_{2}$, shown as green circles, are a triplet of consecutive input frames. If we generate intermediate frames at time $t$ between $(\textbf{I}_{0}, \textbf{I}_{1})$ and between $(\textbf{I}_{1}, \textbf{I}_{2})$, and subsequently generate back an intermediate frame at time $(1-t)$ between $(\hat{\textbf{I}}_{t}, \hat{\textbf{I}}_{t+1})$, the resulting frame must match the original middle input frame, $\textbf{I}_{1}$.}
\label{fig:cycle_consistency_figure}
\end{figure}

We then optimize $\mathcal{M}$ to minimize the reconstruction error between $\hat{\textbf{I}}_{1}$ and $\textbf{I}_{1}$, as
\begin{equation} \label{eq:3}
\operatorname*{arg\,min}_{ \bold{\theta}^ { (\mathcal{M}) }  }   \Big(
\big\lVert \hat{\textbf{I}}_{1} - \textbf{I}_{1} \big\rVert_1 \Big).
\end{equation}
\noindent Figure \ref{fig:cycle_consistency_figure} schematically presents our cycle consistency based approach.

A degenerate solution to optimizing equation \ref{eq:3} might be to copy the input frames as the intermediate predictions (i.e. outputs). However, in practice this does not occur. In order for $\mathcal{M}$ to learn to do copy frames in this way, it would have to learn to identify the input's time information within a single forward operation (eq.~\ref{eq:2}), as $\textbf{I}_{1}$ is a $t=1$ input in the first pass, and $\textbf{I}_{1}$ is a $t=0$ input in the second pass. This is difficult, since the same weights of $\mathcal{M}$ are used in both passes. We support this claim in all of our experiments, where we compared our learned approach using equation~\ref{eq:3} with the trivial case of using inputs as intermediate prediction.

It is true that triplets of input frames could be exploited directly. For example, the reconstruction error between $\mathcal{M}(\textbf{I}_{0},\textbf{I}_{2},t=0.5)$ and $\textbf{I}_{1}$ could be used without cycle consistency. However, our experiments in Section \ref{large_step_supervision} suggest that larger time-step lead to significantly worse accuracy if used without cycle consistency. 

Optimizing $\mathcal{M}$ to the satisfy cycle consistency (CC) constraint in time, as will show in our experiments in Sections  \ref{Unsupervised_large_scale_training} and \ref{domain_transfer}, is effective and is able to generate arbitrarily many intermediate frames that are realistic and temporally smooth. It also produces results that are competitive with supervised approaches, including the same model $\mathcal{M}$, but trained with supervision.

In this work, we also propose techniques that can make unsupervised fine-tuning processes robust. It is quite common to have access to out-of-domain training videos in abundance or access to already pre-trained interpolation models. On the other hand, target domain videos are often limited in quantity, and most critically, lack ground truth intermediate frames. We aim to optimize $\mathcal{M}$ in target videos to jointly satisfy cycle consistency as defined in equation \ref{eq:3} and also learn to approximate a known pre-trained interpolation model, denoted as $\mathcal{F}$. Mathematically, our modified objective is given as,
\begin{equation} 
\label{eq:4}
\begin{split}
\operatorname*{arg\,min}_{ \bold{\theta}^ { (\mathcal{M}) }  } \Big(
& \big\lVert \hat{\textbf{I}}_{1} - \textbf{I}_{1}\big\rVert_1 + \big\lVert \hat{\textbf{I}}_{t} - \mathcal{F}\big(\textbf{I}_{0},\textbf{I}_{1},t\big) \big\rVert_1 + \\
& \big\lVert \hat{\textbf{I}}_{t+1} - \mathcal{F}\big(\textbf{I}_{1},\textbf{I}_{2},t\big) 
\big\rVert_1
\Big) , 
\end{split}
\end{equation}
where $\hat{\textbf{I}}_{1}$ is the cycle reconstructed frame given by equation \ref{eq:2}, $\hat{\textbf{I}}_{t}$ and $\hat{\textbf{I}}_{t+1}$ are given by equation \ref{eq:1}, and  $\bold{\theta}^{(\mathcal{M})}$ are the parameters of $\mathcal{M}$ that our optimization processes update. 

The added objective function to approximate $\mathcal{F}$, help regularize $\mathcal{M}$ to generate realistic \textit{hidden} intermediate frames $\hat{\textbf{I}}_{t}$ an $\hat{\textbf{I}}_{t+1}$ by constraining them to resemble predictions of a known frame interpolation model, $\mathcal{F}$. As optimization progresses and $\mathcal{M}$ learns to pick-up interpolation concepts, one can limit the contribution of the regularizing ``pseudo'' supervised (PS) loss and let optimizations be guided more by the cycle consistency. Such a surrogate loss term, derived from estimated intermediate frames, can make our training processes converge faster or make our optimization processes robust by exposing it to many variations of $\mathcal{F}$.

In this work, for the sake of simplicity, we chose $\mathcal{F}$ to be the same as our $\mathcal{M}$, but pre-trained with supervision on a disjoint dataset that has ground-truth high frame rate video, and denote it as $\mathcal{M}_{pre}$. Our final objective can be given by,
\begin{equation} 
\label{eq:5}
\begin{split}
\operatorname*{arg\,min}_{ \bold{\theta}^ { (\mathcal{M}) }  }   \Big(
&\lambda_{rc}\big\lVert \hat{\textbf{I}}_{1} - \textbf{I}_{1} \big\rVert_1  + \lambda_{rp}\big\lVert \hat{\textbf{I}}_{t} - \mathcal{M}_{pre}\big(\textbf{I}_{0},\textbf{I}_{1},t\big) \big\rVert_1 + \\
&\lambda_{rp}\big\lVert \hat{\textbf{I}}_{t+1} - \mathcal{M}_{pre}\big(\textbf{I}_{1},\textbf{I}_{2},t\big) 
\big\rVert_1
\Big), 
\end{split}
\end{equation}
\noindent where $\lambda_{rc}$ and $\lambda_{rp}$ are weights of CC and PS losses.

As we will show in the experiments, optimizing equation \ref{eq:5} by relying only on the PS loss, without cycle consistency, will teach $\mathcal{M}$ to perform at best as good as $\mathcal{M}_{pre}$, i.e., the model used in the same PS loss. However, as we show in Section~\ref{optimal_PSL_weight}, by weighting cycle consistency and PS losses appropriately, we achieve frame interpolation results that are superior to those obtained by learning using either CC or PS losses alone.

Finally, we implement our $\mathcal{M}$ using the Super SloMo video interpolation model \cite{jiang2018super}. Super SloMo is a state of the art flow-based CNN for video interpolation, capable of synthesizing an arbitrary number of high quality and temporally stable intermediate frames. Our technique is not limited to this particular interpolation model, but could be adopted with others as well.

In the subsequent subsections we provide a short summary of the Super SloMo model, our loss functions, and techniques we employed to make our unsupervised training processes stable.

\subsection{Video Interpolation Model}
To generate one or more intermediate frames $\hat{\textbf{I}}_{t}$ from a pair of input frames $(\textbf{I}_{0}, \textbf{I}_{1})$, first the Super SloMo model estimates an approximate bi-directional optical flow from any arbitrary time $t$ to $0$, $\textbf{F}_{t\rightarrow0}$, and from $t$ to $1$, $\textbf{F}_{t\rightarrow1}$. Then, it generates a frame by linearly blending the input frames after they are warped by the respective estimated optical flows, as
\begin{equation} \label{eq:6}
\hat{\textbf{I}}_{t} = \alpha \mathcal{T}(\textbf{I}_{0},\textbf{F}_{t\rightarrow0}) + (1-\alpha) \mathcal{T}(\textbf{I}_{1},\textbf{F}_{t\rightarrow1}) , 
\end{equation}

\noindent where $\mathcal{T}$ is an operation that bilinearly samples input frames using the optical flows, and $\alpha$ weighs the contribution of each term. The blending weight $\alpha$ models both global property of temporal consistency as well as local or pixel-wise occlusion or dis-occlusion reasoning. For instance, to maintain temporal consistency, $\textbf{I}_{0}$ must contribute more to $\hat{\textbf{I}}_{t}$ when $t$ is close to $0$. Similarly, $\textbf{I}_{1}$ contributes more to $\hat{\textbf{I}}_{t}$, when $t$ is close to $1$. 

To cleanly blend the two images, an important property of video frame interpolation is exploited, i.e. not all pixels at time $t$ are visible in both input frames. Equation \ref{eq:6} can thus be defined by decomposing $\alpha$ to model both temporal consistency and occlusion or de-occlusions, as
\begin{equation}  \label{eq:7}
\hat{\textbf{I}}_{t} =\frac{1}{Z}\Big(\big(1-t\big)\textbf{V}_{t\leftarrow0} \mathcal{T}\big(\textbf{I}_{0},\textbf{F}_{t\rightarrow0}\big) + t\textbf{V}_{t\leftarrow1} \mathcal{T}\big(\textbf{I}_{1},\textbf{F}_{t\rightarrow1}\big)\Big) ,
\end{equation}
\noindent where $\textbf{V}_{t\leftarrow0}$ and $\textbf{V}_{t\leftarrow0}$ are visibility maps, and $Z=(1-t)\textbf{V}_{t\leftarrow0}+t\textbf{V}_{t\leftarrow1}$ is a normalisation factor. For a pixel $p$,  $\textbf{V}_{t\leftarrow0}(p) \in [0,1]$ denotes visibility of $p$ at time $t$ ($0$ means fully occluded or is invisible at $t$). 

The remaining challenge is estimating the intermediate bi-direction optical flows $(\textbf{F}_{t\rightarrow0}, \textbf{F}_{t\rightarrow1})$ and the corresponding visibility maps $(\textbf{V}_{t\leftarrow0}, \textbf{V}_{t\leftarrow1})$. For more information, we refer the reader to \cite{jiang2018super}.

\subsection{Training and Loss Functions}
We train $\mathcal{M}$ to generate arbitrarily many intermediate frames $\{{\hat{\textbf{I}}_{t_{i}}}\}_{i=1}^{N}$ without using the corresponding ground-truth intermediate frames $\{{\textbf{I}_{t_{i}}}\}_{i=1}^{N}$, with $N$ and $t_{i}\in(0,1)$ being frame count and time, respectively. Specifically, as described in Section \ref{Methodology}, we optimize $\mathcal{M}$ to (a) minimize the errors between the cycle reconstructed frame $\hat{\textbf{I}}_{1}$ and $\textbf{I}_{1}$ and (b) to minimize the errors between the intermediately predicted frames $\hat{\textbf{I}}_{t}$ and $\hat{\textbf{I}}_{t+1}$ and the corresponding estimated or pseudo ground-truth frames $\mathcal{M}_{pre}(\textbf{I}_{0},\textbf{I}_{1},t)$ and $\mathcal{M}_{pre}(\textbf{I}_{1},\textbf{I}_{2},t)$. 

Note that, during optimization a cycle reconstructed frame $\hat{\textbf{I}}_{1}$ can be obtained via arbitrarily many intermediately generated frames $\{\hat{\textbf{I}}_{t_{i}},\hat{\textbf{I}}_{t_{i}+1}\}_{i=1}^{N}$. Thus, many reconstruction errors can be computed from a single triplets of training frames $\{\textbf{I}_{0},\textbf{I}_{1},\textbf{I}_{2}\}$. However, we found doing so makes  optimizations unstable and often unable to converge to acceptable solutions. Instead, we found establishing very few reconstruction errors per triplet to make our training stable and generate realistic intermediate frames. In our experiments, we calculate one reconstruction error per triplet, at random time $t_{i}\in(0,1)$. 

Our training loss functions are given by, 
\begin{equation}\label{eq:8}
\mathcal{L} = \lambda_{rc}\mathcal{L}_{rc} + \lambda_{rp}\mathcal{L}_{rp} + \lambda_{p}\mathcal{L}_{p} + \lambda_{w}\mathcal{L}_{w} + \lambda_{s}\mathcal{L}_{s} , 
\end{equation}
where $\mathcal{L}_{rc}$, defined as, 
\begin{equation}\label{eq:9}
\mathcal{L}_{rc} = {\lVert\hat{\textbf{I}}_{1}-\textbf{I}_{1}\rVert}_{1} , 
\end{equation}
models how good the cycle reconstructed frame is, and $\mathcal{L}_{rp}$, defined as, 

\begin{equation}\label{eq:10}
\begin{split}
\mathcal{L}_{rp} = &{\lVert\hat{\textbf{I}}_{t_{i}}-\mathcal{M}_{pre}(\textbf{I}_{0},\textbf{I}_{1},t_{i})\rVert}_{1} + \\
&{\lVert\hat{\textbf{I}}_{t_{i}+1}-\mathcal{M}_{pre}(\textbf{I}_{1},\textbf{I}_{2},t_{i})\rVert}_{1}, 
\end{split}
\end{equation}

 models how close the \textit{hidden} intermediate frames are to our pseudo intermediate frames. $\mathcal{L}_{p}$ models a perceptual loss defined as the $L_{2}$ norm on the high-level features of VGG-16 model, pre-trained on ImageNet, and is given as, 
\begin{equation}\label{eq:11}
\mathcal{L}_{p} = {\lVert\Psi(\hat{\textbf{I}}_{1})-\Psi(\textbf{I}_{1})\rVert}_{2} , 
\end{equation}
with $\Psi$ representing the \texttt{conv4\_3} feature of the VGG-16 model.

Our third loss, $\mathcal{L}_{w}$ is a warping loss that make optical flow predictions realistic, and is given by, 
\begin{equation}\label{eq:12}
\begin{split}
\mathcal{L}_{w} = & 
{\lVert\mathcal{T}(\textbf{I}_{0}, \textbf{F}_{1\rightarrow 0})-\textbf{I}_{1}\rVert}_{1} + {\lVert\mathcal{T}(\textbf{I}_{1}, \textbf{F}_{0\rightarrow 1})-\textbf{I}_{0}\rVert}_{1} + \\
& {\lVert\mathcal{T}(\textbf{I}_{1}, \textbf{F}_{2\rightarrow1})-\textbf{I}_{2}\rVert}_{1} + {\lVert\mathcal{T}(\textbf{I}_{2}, \textbf{F}_{1\rightarrow2})-\textbf{I}_{1}\rVert}_{1} + \\
& {\lVert\mathcal{T}(\hat{\textbf{I}}_{t}, \textbf{F}_{t+1\rightarrow t})-\hat{\textbf{I}}_{t+1}\rVert}_{1} + {\lVert\mathcal{T}(\hat{\textbf{I}}_{t+1}, \textbf{F}_{t\rightarrow t+1})-\hat{\textbf{I}}_{t}\rVert}_{1}.
\end{split}
\end{equation}

In a similar way as the Super SloMo framework, we also enforce a smoothness constraint to encourage neighbouring optical flows to have similar optical flow values, and it is given as, 
\begin{equation}\label{eq:13}
\begin{split}
\mathcal{L}_{s} = & {\lVert\Delta\textbf{F}_{t\rightarrow t+1}\rVert}_{1} + {\lVert\Delta\textbf{F}_{t+1\rightarrow t}\rVert}_{1} + \\
& {\lVert\Delta\textbf{F}_{0\rightarrow1}\rVert}_{1} + {\lVert\Delta\textbf{F}_{1\rightarrow0}\rVert}_{1} + \\
& {\lVert\Delta\textbf{F}_{1\rightarrow2}\rVert}_{1} + {\lVert\Delta\textbf{F}_{2\rightarrow1}\rVert}_{1} ,
\end{split}
\end{equation}
where $\textbf{F}_{t\rightarrow t+1}$ and $\textbf{F}_{t+1\rightarrow t}$ are the forward and backward optical flows between the the intermediately predicted $\hat{\textbf{I}}_{t}$ and $\hat{\textbf{I}}_{t+1}$ frames.

Finally, we linearly combine our losses using experimentally selected weights: $\lambda_{rc}=0.8$, $\lambda_{rp}=0.8$, $\lambda_{p} = 0.05$, $\lambda_{w} = 0.4$, and $\lambda_{s}=1$, see Section \ref{optimal_PSL_weight} for details on weight selection. 

\begin{table*}[ht!]
\small
\centering
\begin{tabular}{c  c  c  c  c c c}
\hline
& FPS & Frame count & Clip count & Resolution & Train & Test \\ 
\hline
UCF101~\cite{soomro2012dataset}  & 25 & 1,137 & 379 & $256\times256$ & & x \\  
YouTube~\cite{jiang2018super} & 240 & 296,352 & 1,014 & $720\times1280$ & x &  \\  
Battlefield-1~\cite{reda2018sdc} & 30 & 329,222 & 363 & $1080\times1920$ & x & \\
\hline
\multirow{2}*{Adobe~\cite{su2017deep} }& 30 & 9,551 & \multirow{2}*{112} & \multirow{2}*{$720\times1280$} & x & \\  
 & 240 & 76,768 &  &  & x & \\  
\hline
\multirow{2}*{Slowflow~\cite{janai2017slow}} & 30 & 3,470 & \multirow{2}*{46} & \multirow{2}*{$2048\times2560$} & x &  \\
 & 240 & 414 &  &  &  & x \\  
\hline 
\multirow{2}*{Sintel~\cite{janai2017slow}} & 24 & 551 & \multirow{2}*{13} & \multirow{2}*{$872\times2048$} & x &  \\
 & 1008 & 559 &  &  &  & x \\  
\hline
\end{tabular}
\caption{Statistics of video datasets used in training or evaluation.}
\label{table:statistics_of_datasets}
\end{table*}

\subsection{Implementation Details}
We use Adam solver \cite{kingma2014adam} for optimization with ${\beta}_{1}=0.9$, ${\beta}_{2}=0.999$, and no weight decay. We train our models for a total of 500 epochs, with a total batch size of 32 on 16 V100 GPUs, using distributed training over two nodes. Initial learning rate is set to $1e^{-4}$, and then scaled-down by 10 after 250, and again after 450 epochs.

\section{Experiments}\label{Results}

\subsection{Datasets and Metrics}

Table \ref{table:statistics_of_datasets} summarizes datasets we used for training and evaluation. We used Adobe-240fps~\cite{su2017deep} (76.7K frames) and YouTube-240fps~\cite{jiang2018super} (296.4K frames) for supervised training to establish baselines. For unsupervised training, we considered low FPS Battlefield1-30fps videos~\cite{reda2018sdc} (320K frames), and Adobe-30fps (9.5K frames), obtained by temporally sub-sampling Adobe-240fps videos, by keeping only every other 8th frame. We chose game frames because they contain a large range of motion that could make learning processes robust. We used UCF101~\cite{soomro2012dataset} datasets for evaluation.

To study our unsupervised fine-tuning techniques in bridging domain gaps, we considered two particularly distinct, high FPS and high resolution, target video datasets: Slowflow-240fps and Sintel-1008fps~\cite{janai2017slow}. Slowflow is captured from real life using professional high speed cameras, whereas Sintel is a game content. We split Slowflow dataset into disjoint low FPS train (3.4K frames) and a high FPS test (414 frames) subsets, see Table \ref{table:statistics_of_datasets}. We create the test set by selecting nine frames in each of the 46 clips. We then create our low FPS train subset by temporally sub-sampling the remaining frames from 240-fps to 30-fps. During evaluation, our models take as input the first and ninth frame in each test clip and interpolate seven intermediate frames. We follow a similar procedure for Sintel-1008fps~\cite{janai2017slow}, but interpolate 41 intermediate frames, i.e., conversion of frame rate from 24- to 1008-fps.

To quantitatively evaluate interpolations we considered Peak-Signal-To-Noise (PSNR), the Structural-Similarity-Image-Metric (SSIM), and the Interpolation-Error (IE) \cite{baker2011database}, which is calculated as the root mean-squared-error between generated and ground truth frames. High PSNR and SSIM scores indicate better quality, whereas for IE score it is the opposite.

\begin{table}[h!]
\small
\centering
\begin{tabular}{l c c c c}
\hline
&  & \multicolumn{3}{c}{\textbf{UCF101}} \\
\hline
 & Training data & PSNR($\uparrow$) & SSIM($\uparrow$) & IE($\downarrow$) \\ 
\hline
Trivial Copy  & N/A & 31.27  & 0.895  & 8.35 \\
Baseline & Adobe-240fps & 34.63 & 0.946 & 5.48  \\  
\multirow{2}*{Proposed} & Adobe-30fps   & 34.47 & 0.946 & 5.50 \\
& BattleField-30fps & 34.55 & 0.947 & 5.38 \\
\hline
\end{tabular}
\caption{Interpolation results for single intermediate frame interpolation on UCF101.}
\label{table:UCF101_evaluation}
\end{table}

\subsection{Large-Scale Unsupervised Training}\label{Unsupervised_large_scale_training}
In this section, we consider the scenario where we do not have any high frame rate videos to train a base model, but we have abundant low frame rate videos.
We test our models on UCF101 dataset; for every triplet of frames, the first and third ones are used as input to predict the second frame.

Results are presented in Table~\ref{table:UCF101_evaluation}.
Our unsupervised technique trained on Adobe-30fps performs competitively with results obtained with supervision on Adobe-240fps, achieving PSNR of 34.47, and 34.63 respectively.
Compared to the supervised training, our unsupervised training uses 1/8th of the frame count, and performs comparably to techniques trained with supervision. This shows the effectiveness of cycle consistency constraint alone in training models, from random initialization, for video frame interpolation.
We further study the impact of frame count in unsupervised training. For this study, we used the low FPS Battlefield-1 sequences. The higher the frame count of low FPS frames, the better our unsupervised model performs, when evaluated on UCF101. Using Battlefield-30fps, at frame count four times larger than Adobe-240fps, we achieve results on par with supervised techniques, achieving IE of 5.38 and 5.48, respectively.

Table~\ref{table:UCF101_evaluation} also presents results of trivial copy, which is the simple case of using inputs as predictions. Compared to cycle consistency, trivial prediction leads to significantly worse interpolation, further indicating our approach does in fact allow us to synthesize intermediate frames from unpaired raw video frames.

\subsection{Unsupervised Fine-tuning for Domain Transfer}\label{domain_transfer}

One particularly common situation in video frame interpolation is that we have access to pre-trained models or access to high FPS out-of-domain videos in abundance, but lack ground truth frames in target videos, which are also commonly limited in quantity. Our unsupervised techniques allow us to fine-tune pre-trained models directly on target videos without additional data, and demonstrate significant gain in accuracy in upscaling frame rates of target videos.

First, we consider the scenario where train and test videos are collected with different camera set-ups. We assume we have access to high fps videos collected by hand-held cameras, which is the Adobe-240fps, YouTube-240fps, UCF101 datasets, and consider the Slowflow dataset as our target, a particularly high resolution and high FPS video captured by high speed professional cameras in real life. 
Our baseline is a frame interpolation model trained with supervision. Specifically, we consider SuperSloMo and Deep Voxel Flow (DVF)~\cite{liu2017video}. DVF is another widely-used single-frame interpolation method. We apply our unsupervised fine-tuning directly on the low FPS train split of Slowflow, and evaluate on its test split.
\begin{table}[h!]
\small
\centering
\begin{tabular}{c c c  c  c  }
\hline
 \multicolumn{5}{c}{\textbf{Adobe}\textrightarrow \textbf{Slowflow}} \\ 
\hline
 & Loss & PSNR($\uparrow$)  & SSIM($\uparrow$) & IE($\downarrow$) \\ 
\hline
Trivial Copy & N/A & 25.00  & 0.718  & 14.86 \\
Baseline & PairedGT & 32.84  & 0.887  & 6.67  \\
\multirow{2}*{Proposed} & CC & 32.35  & 0.886  & 6.78  \\ 
 & CC + PS & \textbf{33.05}  & \textbf{0.890}  & \textbf{6.62}  \\ 
\hline
\multicolumn{5}{c}{\textbf{Adobe+YouTube}\textrightarrow \textbf{Slowflow}} \\
 \hline
Baseline & PairedGT & 33.13  & 0.889  & 6.63 \\
Proposed &  CC + PS & \textbf{33.20}  & \textbf{0.891}  & \textbf{6.56} \\
  \hline
\end{tabular}
\caption{Multi-frame interpolation results on Slowflow for frame rate conversion from 30- to 240-FPS, and domain transfer experiments using baselines obtained by pre-training with supervision on Adobe- or Adobe+YouTube-240FPS.}
\label{table:Slowflow_domain_transfer}
\end{table}

 \begin{figure}[t]
    \centering
    \includegraphics[trim={0 0 0 0},clip,width=1.0\linewidth]{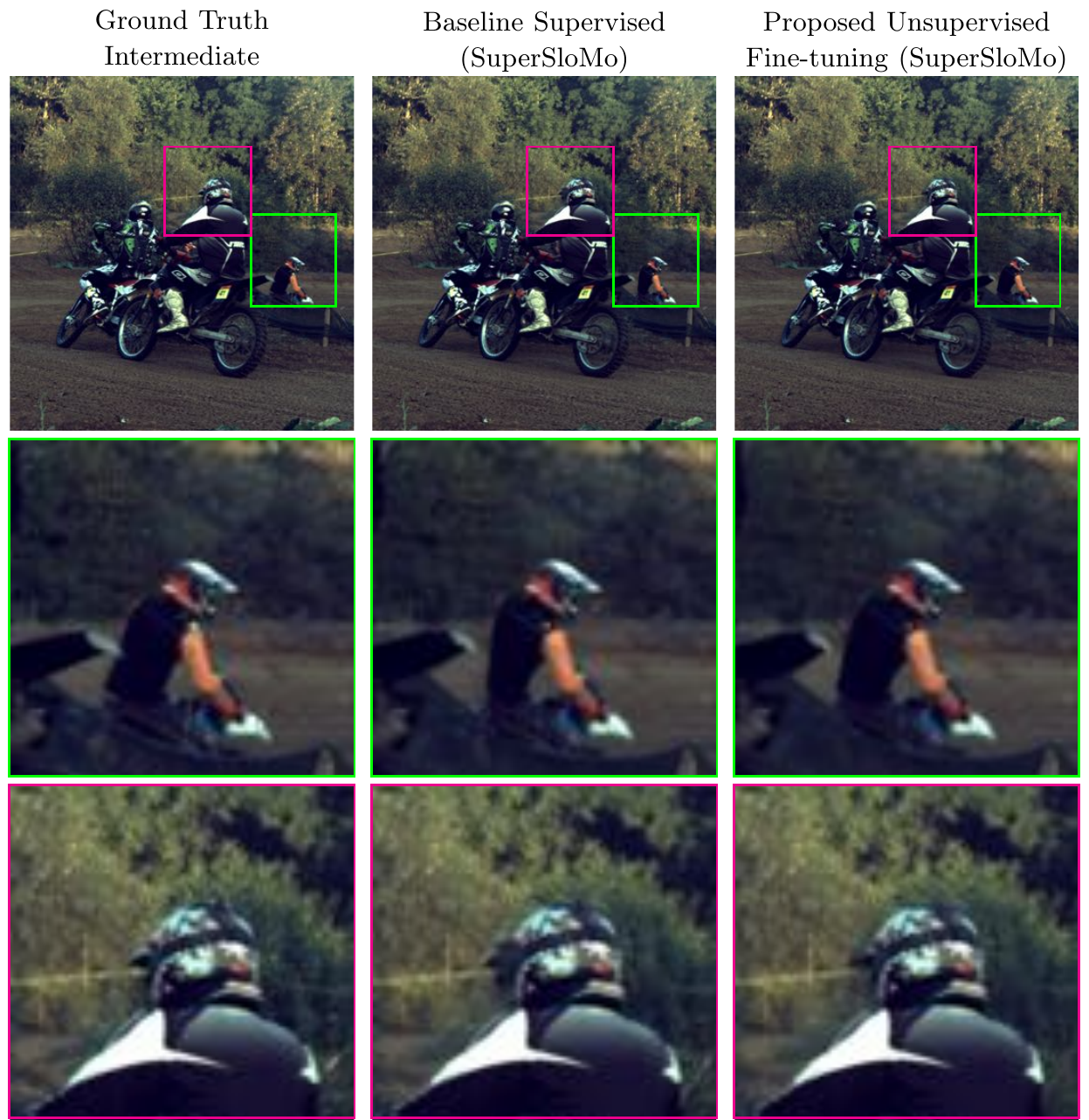}
    \caption{Visual results of a sample from Slowflow dataset. Baseline supervised model is trained with Adobe dataset and proposed unsupervised model is fine-tuned with Slowflow. Improvements seen as the person's back squeezed in supervised (middle) but preserved in unsupervised (right). On bottom row, although both techniques blur the regions surrounding the helmet, the shape of helmet is preserved in our proposed technique. 
    }
\label{fig:Sf_domain_transfer}
\end{figure}

\begin{table}[h!]
\small
\centering
\begin{tabular}{c  c c  c  c }
\hline
 \multicolumn{5}{c}{\textbf{Adobe}\textrightarrow \textbf{Sintel}} \\
\hline
 & Loss & PSNR($\uparrow$)  & SSIM($\uparrow$) & IE($\downarrow$)  \\ 
\hline
Trivial Copy  & N/A & 22.48  & 0.714  & 20.23 \\
Baseline & PairedGT & 31.82 & 0.912 & 5.61 \\
\multirow{2}*{Proposed} & CC & 30.08 & 0.872 & 7.67 \\ 
 & CC+PS & \textbf{32.53} & \textbf{0.918}& \textbf{5.21}  \\ 
\hline
\multicolumn{5}{c}{\textbf{Adobe+YouTube}\textrightarrow \textbf{Sintel}} \\
\hline
Baseline & PairedGT  & 33.23 & 0.928 & 4.74 \\
{Proposed}& CC+PS &  \textbf{33.34} & \textbf{0.928}& \textbf{4.71} \\
 \hline
\end{tabular}
\caption{Multi-frame interpolation results on Sintel for frame rate conversion from 24 to 1008 FPS, and domain transfer experiments using baselines obtained by pre-training with supervision on Adobe- or Adobe+YouTube-240fps.}
\label{table:Sintel_domain_transfer}
\end{table}

\noindent\textbf{Adobe\textrightarrow Slowflow}: Our unsupervised training with cycle consistency alone performs quite closely to the baseline (Super SloMo pre-trained with supervision), achieving PSNR of 32.35 and 32.84, respectively. While a total of 76.7K Adobe-240fps frames are used in supervised pre-training, our unsupervised training is performed with only 3K frames of Slowflow, which indicates the efficiency and robustness of our proposed unsupervised training technique. Furthermore, fine-tuning the pre-trained model by jointly optimizing to satisfy cycle consistency and to minimize our proposed pseudo supervised loss (CC + PS), we outperform the pre-trained baseline by a large margin, with PSNR of 33.05 vs. 32.84. The PS loss relies on the same pre-trained baseline model, as discussed in Section \ref{Methodology}, and regularizes our training process. If used alone, i.e without cycle consistency, it performs at best as good as the baseline pre-trained model, see Section \ref{optimal_PSL_weight} for more details.

\noindent\textbf{Adobe+YouTube\textrightarrow Slowflow}: Here, our baseline model is pre-trained on a larger dataset Adobe+YouTube, total of 372.7K frames, and achieves better accuracy than pre-training on Adobe alone, achieving PSNR 33.13 vs. 32.84, when directly applied on Slowflow test videos. Even with improved pre-trained baseline, we observe consistent benefit with our proposed unsupervised fine-tuning, improving PSNR from 33.13 to 33.20.

Another interesting observation from this study is that it takes an extra 296.K frames from YouTube-240fps to improve PSNR from 32.84 to 33.13 on Slowflow, via pre-training with supervision. We achieve a comparable improvement of PSNR from 32.84 to 33.05 by simply fine-tuning on the target low FPS frames in a completely unsupervised way. 
Sample interpolation results from this study can be found at Figure \ref{fig:Front_page_figure_slowflow}, where improvements on the bicycle tire and the shoe are highlighted, and at Figure \ref{fig:Sf_domain_transfer}, where improvements on the persons' back and helmet regions are highlighted.

Table \ref{table:Sintel_domain_transfer} present results of unsupervised fine-tuning for domain transfer from Adobe\textrightarrow Sintel and Adobe+YouTube\textrightarrow Sintel for the task of upscaling frame rate from 24- to 1008-fps. Similarly to the Slowflow experiments, in Table \ref{table:Slowflow_domain_transfer}, the results indicate the utility of our unsupervised techniques in shrinking domain gaps or achieving results that compete with supervised techniques.

\begin{table}[h]
\small
\renewcommand\tabcolsep{3.0pt}
\centering
\begin{tabular}{c  c  c  c   |  c  c  c  c}
\hline
& \multicolumn{3}{c}{\textbf{Slowflow}} & \multicolumn{3}{c}{\textbf{Sintel}} \\
\hline
 & PSNR($\uparrow$)  & SSIM($\uparrow$) & IE($\downarrow$) &  PSNR($\uparrow$)  & SSIM($\uparrow$) & IE($\downarrow$) \\ 
\hline
Baseline & 30.79 & 0.84 & 8.57 & 29.14  & 0.84  & 8.96\\ 
Proposed & \textbf{31.42} & \textbf{0.86}& \textbf{7.99} & \textbf{29.71}  & \textbf{0.86}  & \textbf{8.23} \\ 
\hline
\end{tabular}
\caption{Comparison of supervised training at quarter resolution (baseline) and unsupervised fine-tuning at full resolution (proposed) for frame rate upscaling from 30 to 240 FPS (Slowflow) and 24 to 1008 FPS (Sintel).}
\label{table:Slowflow_Sintel_spatial_transfer}
\end{table}

\begin{figure}[h!]
\begin{center}
   \includegraphics[width=1.0\linewidth, trim={0 6.4cm 0 0}, clip]{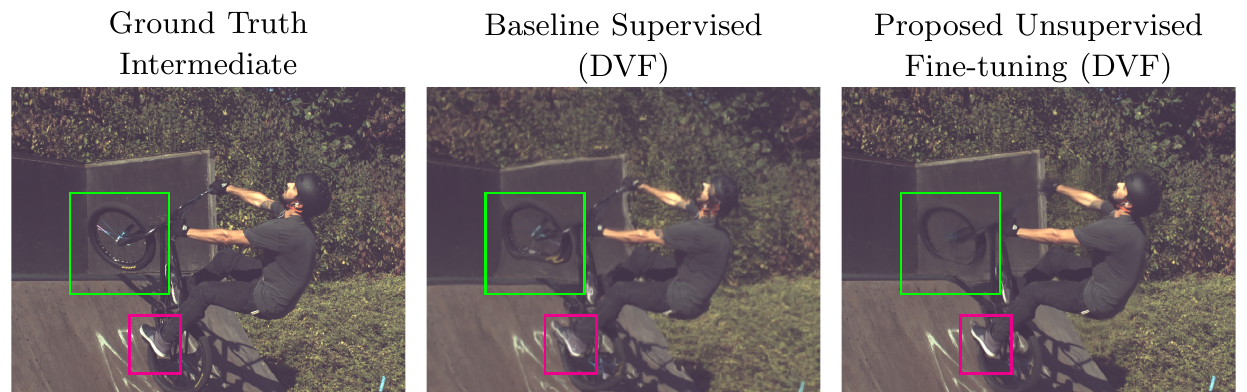}
\end{center}
\vspace{-0.5cm}
   \caption{Visual comparison of unsupervised fine-tuning of DVF with supervised DVF pre-trained on UCF-101.}
\label{fig:Slowflow_domain_transfer_for_DVF}
\end{figure}

\noindent\textbf{UCF101\textrightarrow Slowflow}: Table~\ref{table:Slowflow_domain_transfer_for_DVF} and Figure~\ref{fig:Slowflow_domain_transfer_for_DVF} present results of fine-tuning DVF on Slowflow. We use an off-the-shelf implementation of DVF, pre-trained on UCF101\footnote{https://github.com/lxx1991/pytorch-voxel-flow}. Our unsupervised techniques improve the PSNR from 24.64dB to 28.38dB, demonstrating that our method generalizes well to different interpolation techniques, and is not limited to SuperSloMo.
\begin{table}[h!]
\small
\centering
\begin{tabular}{c c c  c  c  }
\hline
 \multicolumn{5}{c}{\textbf{UCF101}\textrightarrow \textbf{Slowflow} using DVF~\cite{liu2017video}} \\ 
\hline
 & Loss & PSNR($\uparrow$)  & SSIM($\uparrow$) & IE($\downarrow$) \\ 
\hline
Trivial Copy  & N/A & 24.26  & 0.698  & 15.60 \\
Baseline & PairedGT & 25.64  & 0.778  & 12.77 \\
Proposed &  CC + PS & \textbf{28.38}  & \textbf{0.820}  & \textbf{9.79} \\
  \hline
\end{tabular}
\caption{Slowflow 30- to 60-FPS conversion. The baseline DVF is pre-trained on UCF-101 with supervision.}
\label{table:Slowflow_domain_transfer_for_DVF}
\end{table}

\begin{figure}[t]
    \centering
    \includegraphics[trim={0 0 0 0},clip,width=1.0\linewidth]{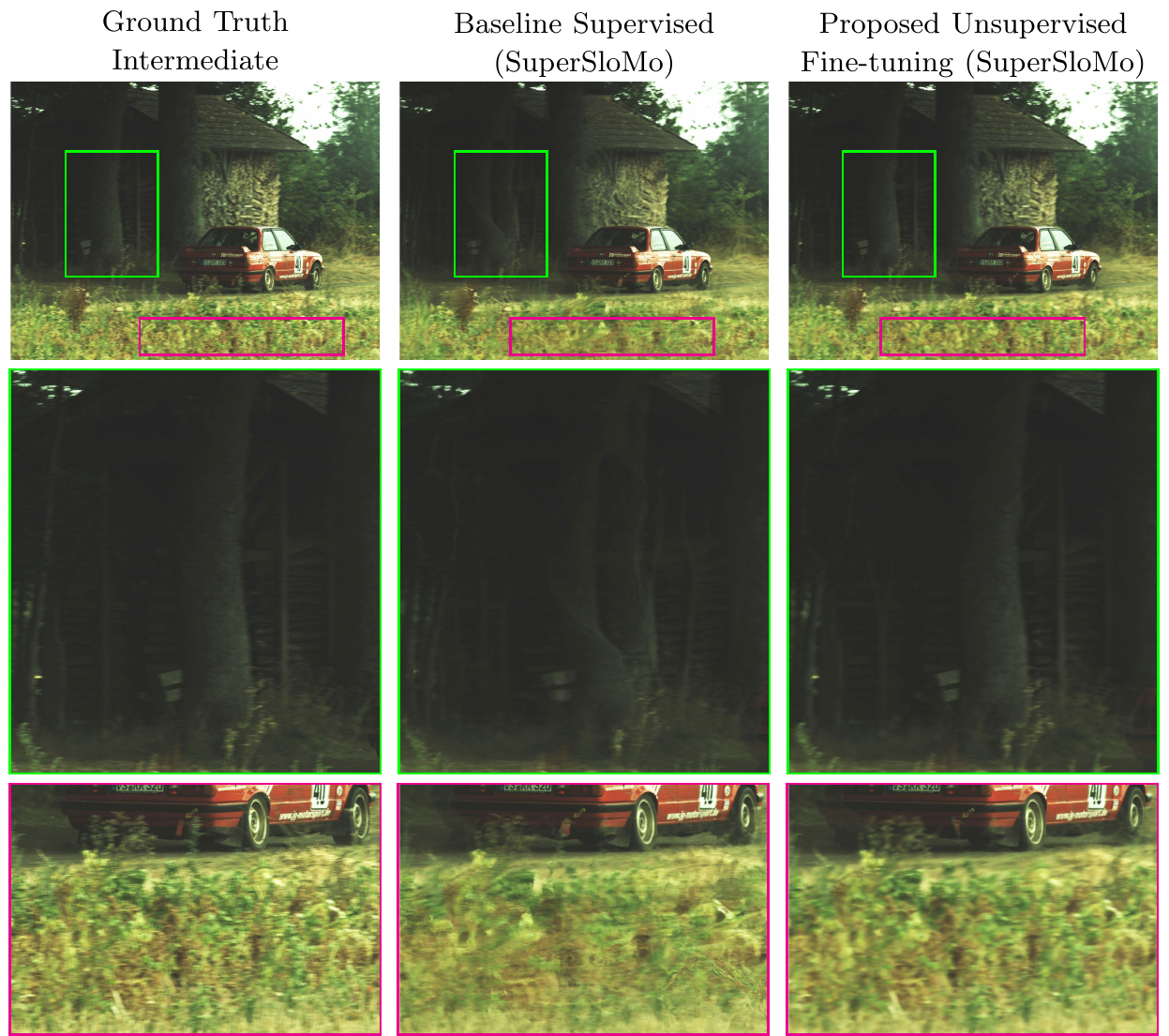}
    \caption{Visual results of a sample from Slowflow dataset. Baseline supervised model is trained with Slowflow dataset in quarter resolution and proposed unsupervised model is fine-tuned with full resolution Slowflow. The tree in the background is deformed, and also deformed in the supervised (middle), while it is predicted well in proposed (right). Bottom row, supervised shows blurriness in the grass, while it is crisper in the proposed.}
    \label{fig:Sf_spatial_transfer}
\end{figure}

In our second domain transfer setting, we consider the scenario where target and test datasets share similarities in content and style but they are in different resolution.
This is a very practical scenario given the scarcity of the high-frame high-resolution videos. Therefore, it is highly desirable to learn from low resolution videos, and be able to interpolate higher resolutions.
We establish a low resolution baseline by training with supervision on  240 fps Slowflow-train dataset, after down-sampling its frames by 4 in each dimension.
Our test video is Slowflow-test split at its original resolution. We repeat similar setting for Sintel. Table~\ref{table:Slowflow_Sintel_spatial_transfer} shows results where our fine-tuning technique on the test domain improves PSNR from 30.79 to 31.42 for Slowflow, and from 29.14 to 29.71 for Sintel.
Visual samples from this experiment can be found in Figure~\ref{fig:Sf_spatial_transfer}.

\subsection{Ablation Studies}

We conduct ablation studies to analyze various design choices of the proposed method. 
\subsubsection{Optimal CC and PS Weights}\label{optimal_PSL_weight}
\begin{figure}[t]
    \centering
    \includegraphics[trim={100 560 100 110},clip,width=1.0\linewidth]{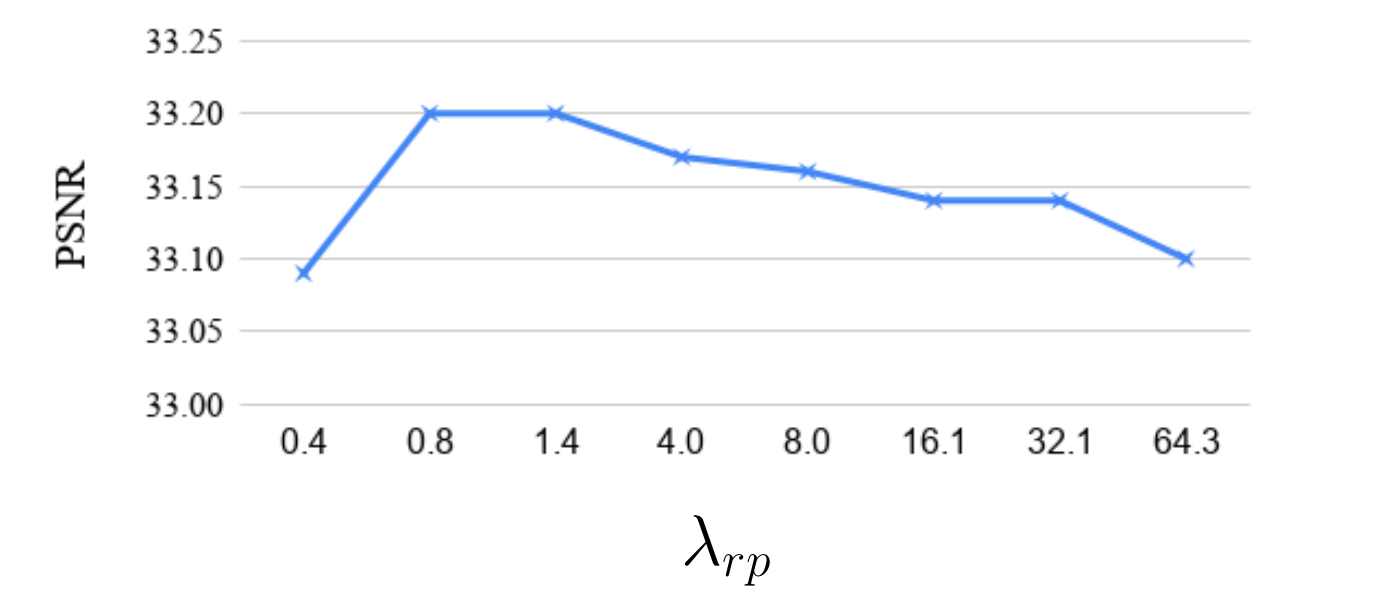}
    \caption{Interpolation accuracy in PSNR versus $\lambda_{rp}$ (PS weight) used in our proposed joint CC and PS optimization techniques, when applied for Adobe+YouTube\textrightarrow Slowflow unsupervised domain transfer. }
    \label{fig:Optimal_PSL_weight_selection}
\end{figure}

Figure \ref{fig:Optimal_PSL_weight_selection} presents interpolation results in PSNR for our models trained with a range of PS weight, $\lambda_{rp}$, values. We fix CC's weight, $\lambda_{rc}$ to 0.8, and vary $\lambda_{rp}$ in small increments from 0 to 64. When $\lambda_{rp}=0$, optimization is guided entirely by CC, it achieves PSNR of 32.35 for unsupervised Adobe+YouTube\textrightarrow Slowflow domain transfer. Interpolation accuracy gradually increases, and plateaus approximately after 0.8. Based on this, we select $\lambda_{rp}=0.8$, and fix its value for all our experiments. At large values of $\lambda_{rp}$, the optimization is mostly guided by PS loss, and as such, trained models perform very similarly to the pre-trained model that the PS loss depends on. Figure \ref{fig:Optimal_PSL_weight_selection} shows this trend. Optimizing with optimally combined CC and PS losses on the other hand leads to results that are superior to those obtained using either loss alone.

\subsubsection{Large Time-step Supervision}
\label{large_step_supervision}
We study the effect of using loss terms, such as $\left \Vert\mathcal{M}(\textbf{I}_{0},\textbf{I}_{2},t=0.5)-\textbf{I}_{1}\right \Vert$ or its variations, defined over a longer time. Table \ref{table:Long_step_objective} presents Adobe+YouTube\textrightarrow Slowflow fine-tuning with cycle consistency, the loss derived from two step interpolation alone or together with cycle consistency. Optimizing using losses derived from long step interpolation result in worse accuracy than optimizing with cycle consistency. When used with cycle consistency, we also did not find it to lead to notable improvement. We attribute this because the model's capacity might be spent to solve the harder problem of interpolating large steps, and provide little benefit to the task of synthesizing intermediate frames between consecutive frames.

\begin{table}[h!]
\small
\centering
\begin{tabular}{c c c  c  c  }
\hline
\multicolumn{5}{c}{\textbf{Adobe+YouTube}\textrightarrow \textbf{Slowflow}} \\ 
\hline
 & Loss & PSNR($\uparrow$)  & SSIM($\uparrow$) & IE($\downarrow$) \\ 
\hline
 & CC & \textbf{32.84}  & \textbf{0.887}  & \textbf{6.67}  \\
 & Long Step & 29.03  & 0.824  & 8.98  \\ 
 & CC + Long Step & 32.24  & 0.884  & 6.81  \\ 
\hline
\end{tabular}
\caption{Comparison of cycle consistency with objectives derived from longer time step interpolation.}
\label{table:Long_step_objective}
\end{table}

\section{Conclusions}
We have presented unsupervised learning techniques to synthesize high frame rate videos directly from low frame rate videos by teaching models to satisfy cycle consistency in time. Models trained with our unsupervised techniques are able to synthesize arbitrarily many, high quality and temporally smooth intermediate frames that compete with supervised approaches. We further apply our techniques to reduce domain gaps in video interpolation by fine-tuning pre-trained models on target videos using a pseudo supervised loss term and demonstrate significant gain in accuracy. Our work shows the potential of learning to interpolate high frame rate videos using only low frame rate videos and opens new avenues to leverage large amounts of low frame rate videos in unsupervised training. 

{\small
\bibliographystyle{ieee_fullname}
\bibliography{egbib}
}


\clearpage
\onecolumn
\section*{Appendix}
\section*{A. Insensitivity to Randomness}
We train our models three times to account for random weight initialisation or random data augmentation. Table \ref{table:PS_loss_only} presents the mean and standard deviation of domain adaptation results for Adobe\textrightarrow Slowflow and Adobe+YouTube\textrightarrow Slowflow. Results indicate our models' insensitivity to randomness, as shown by the margins of improvements in PSNR from 32.84 to 33.05 for Adobe, or from 33.13 to 33.20 for Adobe+YouTube. Table \ref{table:PS_loss_only} also shows fine-tuning with PS loss alone leads to results that are similar to the Baseline.

\section*{B. Interpolation Result vs Intermediate Time}
\begin{figure*}[!b]
    \centering
    \includegraphics[trim={60 120 60  120},clip,width=0.6\linewidth]{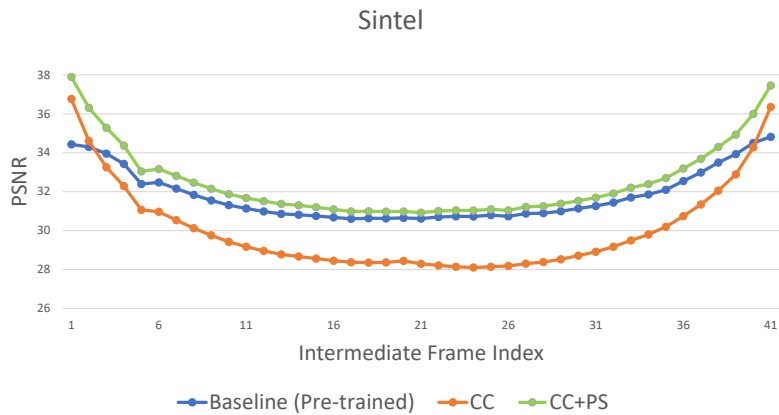}
    \caption{Mean PSNR score at each of the 41 time points for Adobe\textrightarrow Sintel domain adaptation using (a) Super SloMo~\cite{jiang2018super} pre-trained with supervision (Baseline), (b) unsupervised fine-tuning with cycle consistency loss alone (CC), and (c) unsupervised fine-tuning with cycle consistency and pseudo supervised losses (CC+PS). }
\label{fig:Sintel_PSNR_vs_frameIndex}
\end{figure*}

Figure \ref{fig:Sintel_PSNR_vs_frameIndex} presents mean PSNR score at each of the 41 time-points for Adobe\textrightarrow Sintel domain adaptation using (a) Super SloMo~\cite{jiang2018super} pre-trained with supervision (Baseline), (b) unsupervised fine-tuning with cycle consistency loss alone (CC), and (c) unsupervised fine-tuning with cycle consistency and pseudo supervised losses (CC+PS). 

For all models, interpolation accuracy decreases as time-points move away from $t=0$ or $t=1$. Compared to the baseline, our CC-based fine-tuning performs better at the end points (close to $t=0$ or $t=1$), and worse at midway points. On the other hand, our CC+PS-based unsupervised fine-tuning achieves the best of both CC and Baseline, performing better than both CC and Baseline at all time points.

\begin{table*}[h!]
\centering
\begin{tabular}{c c c  c  c  }
\hline
 \multicolumn{5}{c}{\textbf{Adobe}\textrightarrow \textbf{Slowflow}} \\ 
\hline
 & Loss & PSNR  & SSIM & IE \\ 
\hline
Baseline & PairedGT & 32.84  & 0.887  & 6.67  \\
\multirow{3}*{Proposed} & CC & 32.33$\pm$0.028  & 0.886$\pm$0.000  & 6.78$\pm$0.021  \\
& PS & 32.88$\pm$0.006  & 0.887$\pm$0.000  & 6.74$\pm$0.006  \\ 
 & CC + PS & \textbf{33.05}$\pm$\textbf{0.006}  & \textbf{0.890}$\pm$\textbf{0.000}  & \textbf{6.62}$\pm$\textbf{0.000}  \\ 
\hline
\multicolumn{5}{c}{\textbf{Adobe+YouTube}\textrightarrow \textbf{Slowflow}} \\
 \hline
Baseline & PairedGT & 33.13  & 0.889  & 6.63 \\
\multirow{3}*{Proposed} &  PS & 33.14$\pm$0.006  & 0.889$\pm$0.000  & 6.63$\pm$0.006 \\
\multirow{3}*{Proposed} & CC & 32.33$\pm$0.028  & 0.886$\pm$0.000  & 6.78$\pm$0.021  \\
 &  CC + PS & \textbf{33.20}$\pm$\textbf{0.006}  & \textbf{0.891}$\pm${0.001}  & \textbf{6.57}$\pm$\textbf{0.010} \\
  \hline
\end{tabular}
\caption{Mean and Standard deviation of PSNR, SSIM, and IE for domain adaptation of upscaling frame rate from 30- to 240-fps for Adobe\textrightarrow or Adobe+YouTube\textrightarrow Slowflow. CC refers to cycle consistency, and PS pseudo supervised loss.}
\label{table:PS_loss_only}
\end{table*}

\end{document}